\documentclass[runningheads]{llncs}

\usepackage{graphicx}
\usepackage{bbm}
\usepackage{amsmath}
\usepackage{amssymb}
\usepackage{float}
\usepackage{caption}
\usepackage{subcaption}
\usepackage{cite}

\DeclareMathOperator*{\argmax}{arg\,max}

\graphicspath{{images/}}

\title{Leveraging Selective Prediction for Reliable Image Geolocation}

\author{Apostolos Panagiotopoulos \and Giorgos Kordopatis-Zilos \and Symeon Papadopoulos}

\institute{Information Technologies Institute, CERTH, Thessaloniki 60361, Greece \\
\email{\{apanag, georgekordopatis, papadop\}@iti.gr}}

\authorrunning{A. Panagiotopoulos et al.}

\begin{document}
\maketitle

\begin{abstract}
Reliable image geolocation is crucial for several applications, ranging from social media geo-tagging to media verification. State-of-the-art geolocation methods surpass human performance on the task of geolocation estimation from images. However, no method assesses the suitability of an image for this task, which results in unreliable and erroneous estimations for images containing ambiguous or no geolocation clues. In this paper, we define the task of image localizability, i.e. suitability of an image for geolocation, and propose a selective prediction methodology to address the task. In particular, we propose two novel selection functions that leverage the output probability distributions of geolocation models to infer localizability at different scales. Our selection functions are benchmarked against the most widely used selective prediction baselines, outperforming them in all cases. By abstaining from predicting non-localizable images, we improve geolocation accuracy from 27.8\% to 70.5\% at the city-scale, and thus make current geolocation models reliable for real-world applications.

\end{abstract}

\section{Introduction}
\begin{figure*}[t]
    \centering
    \includegraphics[width=\linewidth]{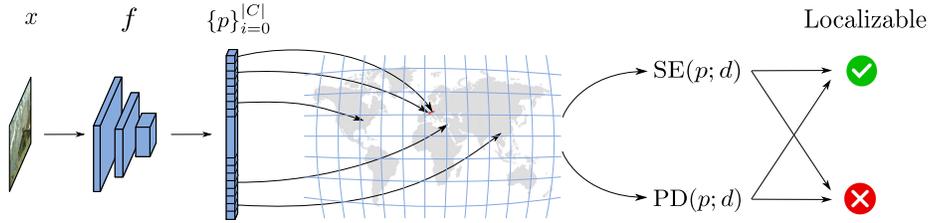}
    \caption{Each image $x$ is passed through a base geolocation model $f$, yielding a cell probability distribution $\{ p \}_{i=1}^{|C|}$.
    Considering this distribution in the context of a world map, we measure whether our model's confidence is concentrated in a particular region or dispersed over globe through our selection functions $\text{SE}(p; d)$ and $\text{PD} (p; d)$, to assess the localizability of the input image.}
    \label{fig:model}
\end{figure*}

A great portion of the images daily uploaded on the Internet are from smartphones and therefore contain geotags, providing information for their geographic location. However, there are numerous cases, such as social media photos, where this information is missing. Therefore, the ability to estimate the geographic location of these images, known as \textit{location estimation} or \textit{geolocation}, is crucial for a number of applications ranging from social media mining to media verification. More formally, image geolocation is the process of inferring the GPS coordinates of the depicted picture based solely on its visual elements.

State-of-the-art approaches for geolocation employ the latest advances in Computer Vision, such as Convolutional Neural Networks (CNNs), to extract representations of the depicted scenes and utilize huge databases of images taken worldwide either for training a classifier \cite{planet, hierarchical, mvmf} or for retrieval \cite{im2gps, large-scale, ranking-fusion, im2gps3k}. Classification solutions partition the earth into a set of geographic cells, and the images are passed through a CNN to be classified into a single cell. Retrieval solutions compare the test images against the ones from a large-scale database in order to retrieve the most similar images and derive a single estimation by aggregating their locations. In both cases, performance is measured by the percentage of images localized within a certain distance $d$ from their ground-truth location, denoted as \textit{geolocation accuracy @ $d$ km}.

Ongoing research focuses mainly on improving geolocation accuracy at different granularities (e.g. $d$ = 1, 25, 200, 750 and 2500 km). However, contrary to most image classification datasets where images usually contain enough visual cues to be classified in a unique class, a good portion of images in geolocation datasets depict  scenes with no apparent visual cues mapping to the image's location (e.g. indoor spaces, portraits). Such images could have been captured anywhere on the globe, and therefore attempting to localize them would most likely result in erroneous or unreliable estimations. Hence, we deem crucial for the reliability of geolocation systems to estimate not only the geolocation of input images but also their \textit{localizability}.

In a general sense, we consider localizable the images that contain enough visual cues for their accurate geolocation. However, localizability is better approached considering a granularity scale (i.e. the range within which an image can be correctly placed from its true location) and a geolocation model (i.e. a mechanism that derives the image's location from its visual content). To this end, we introduce the problem of \textit{image localizability detection}, building upon the foundation of \textit{selective prediction} \cite{foundations}. We propose a methodology that utilizes state-of-the-art geolocation systems to infer localizability at different scales. More precisely, we re-implement an image geolocation model \cite{planet} and, instead of interpreting the output probability distribution as pure categorical data and predicting the most probable location, we visualize the whole distribution on the world map. We then devise two novel selection functions, i.e. spatial entropy and prediction density, which measure cell probability dispersion and concentration over the globe, exploiting intrinsic proprieties of geolocation systems -- unlike current state-of-the-art selection functions. We extensively evaluate our methodology on the two most widely used evaluation datasets, i.e. Im2GPS \cite{im2gps} and Im2GPS3k \cite{im2gps3k}, and highlight the effectiveness of our proposed selection functions compared to state-of-the-art approaches in selective prediction. By discarding images considered non-localizable at city-scale, we boost the accuracy of our base geolocation model at city-scale from 27.8\% to 70.5\%, making it reliable for real-world applications. To the best of our knowledge, we are the first to propose the task of image localizability detection and leverage selective prediction to address it. Therefore, our work makes the following contributions:

\begin{itemize}
    \item We introduce the problem of image localizability detection and frame it under a selective prediction framework. This formulation allows current classification models to infer localizability, and hence abstain from predicting non-localizable images.
    
    \item We propose two novel selection functions, specifically designed for geolocation, that outperform current state-of-the-art selection functions which do not consider spatial information.
    
    \item We extensively evaluate our methodology on the two most widely used datasets, achieving good separation between localizable and non-localizable images, and making current geolocation systems more reliable.
\end{itemize}

\section{Related Work}
This section gives an overview of some of the fundamental works that have contributed to geolocation estimation and selective prediction.

\textbf{Geolocation Estimation:} Hays and Efros \cite{im2gps} introduced the problem of planet-scale image location estimation. They used handcrafted features to retrieve images similar to a query image and infer its location based on theirs. Weyand et al. \cite{planet} took advantage of the deep learning advances and trained a Convolutional Neural Network (CNN) to extract features from images. Additionally, they formulated the geolocation problem as a classification task and divided the earth's surface, using Google's s2 geometry library\footnote{\url{https://s2geometry.io}}, to create a set of classes for the training and test images. More recent works modify the classification pipeline using a hierarchical partitioning of the earth \cite{hierarchical}, or novel loss functions \cite{mvmf}. Recently, a hybrid scheme called Search within Cell \cite{kordopatis-geolocation} was proposed, which combines a classification and retrieval approach for the final location estimation.

\textbf{Selective prediction:} Works in this area focus on machine learning systems that are not only able to make predictions but also to know when to abstain from predicting. Although the field exists for several decades, it was not until recently that a unified formulation was introduced \cite{foundations} and approaches regarding deep architectures were proposed by El-Yaniv and Wiener \cite{selective-prediction, selectivenet}. In \cite{selective-prediction}, Softmax Response (the maximum output after the softmax layer) and  MC-Dropout \cite{mc-dropout} were used as confidence functions, and an algorithm that finds the appropriate threshold given a desired risk was proposed. In \cite{selectivenet}, an algorithm for jointly training the classification network and the selection function was proposed.

\section{Methodology}
\label{sec:methodology}

In this section, we present the proposed methodology for the selection of localizable images; this is illustrated in Fig. \ref{fig:model}.

\subsection{Geolocation Estimation}
\label{sec:geolocation-estimation}
A geolocation model $f$ takes as input an image $x \in \mathbb{R}^{H \times W \times 3}$ and returns the estimated GPS coordinates $\hat{y} \in \mathbb{R}^2$ of the location it was captured. Following the classification approach for geolocation, we first divide the earth's surface into a grid of geographic cells $C$, and then we employ a CNN, with $|C|$ outputs in the final layer, corresponding to the cells of the grid. For each input image $x$, the CNN creates a probability distribution over the grid of cells. The predicted location $\hat{y}$ is the mean coordinates of the cell with the highest probability.

\begin{figure}[t]
    \centering
    
    \begin{subfigure}[h]{0.48\linewidth}
        \includegraphics[width=\linewidth]{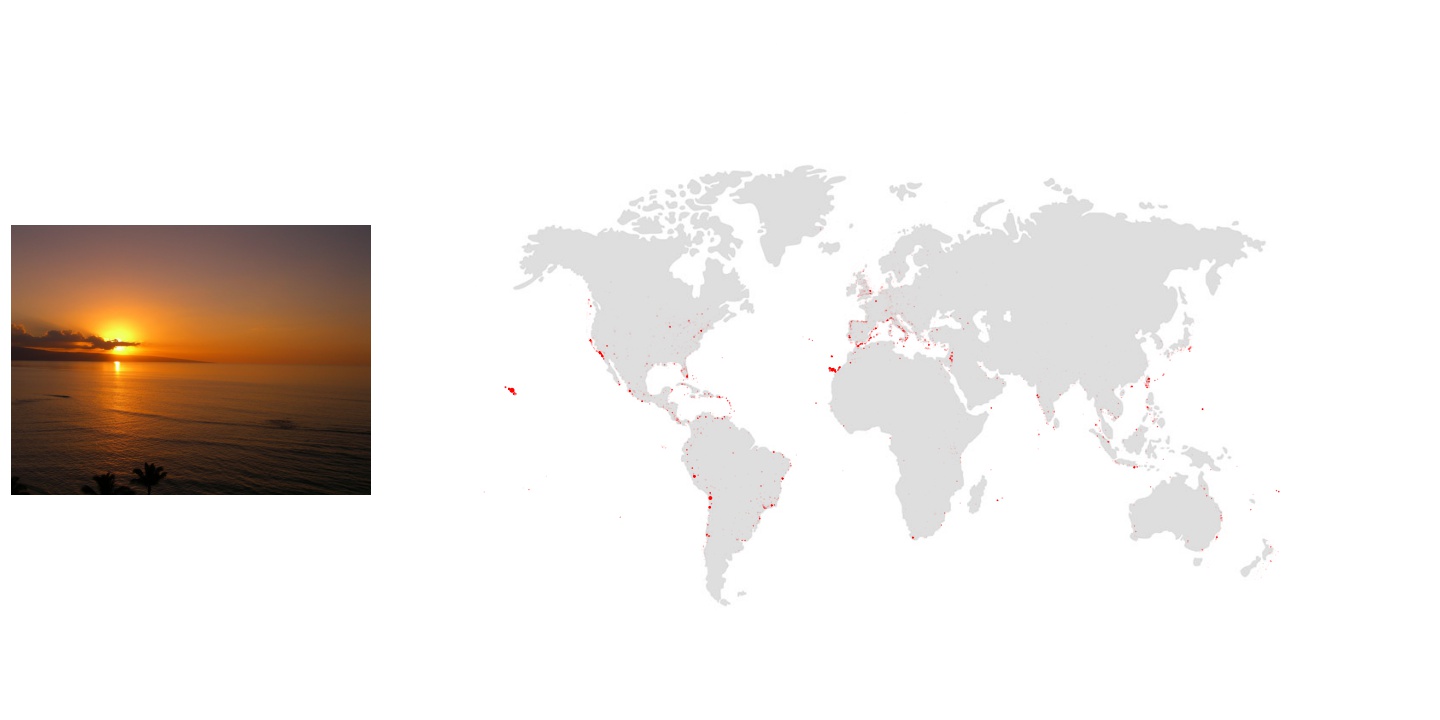}
        \caption{A sunset picture in Hawaii}
        \label{fig:sunset-map}
    \end{subfigure}
    \quad
    \begin{subfigure}[h]{0.48\linewidth}
        \includegraphics[width=\linewidth]{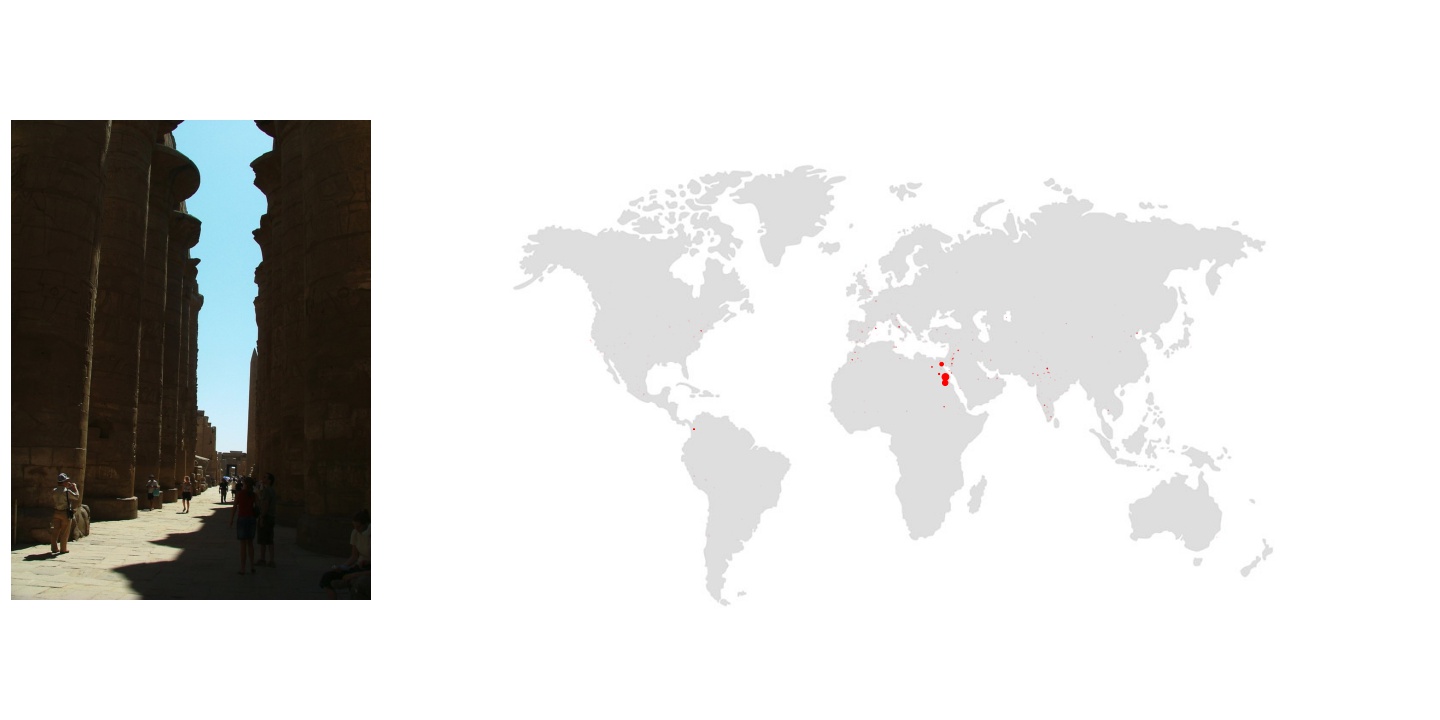}
        \caption{A monument in Egypt}
        \label{fig:egypt-map}
    \end{subfigure}
    
    \caption{Placing our model's cell probability distribution on a map illustrates its ability to associate visual \textit{concepts} to locations.}
    \label{fig:maps}
\end{figure}

Most geolocation approaches \cite{planet, hierarchical, mvmf} consider only the cell with the highest probability, ignoring all the information provided by the cell probability distribution over the grid. We found that this probability distribution can provide valuable insights for the localizability of images. More precisely, a trained network has learned both to estimate the location of images and also to associate \textit{concepts} with locations. For example, when the network is presented with the image of Fig. \ref{fig:sunset-map}, the probability of several cells nearby the sea is high. However, when presented with the image from Fig. \ref{fig:egypt-map}, all cells around Egypt are activated, since the image contains many visual cues that map to that area. Thus, even though it is challenging to predict the exact location of those images, the model generates \textit{reasonable} estimates to candidate locations.

By inspecting the map, it is evident that the network is more confident for the estimation of Fig. \ref{fig:egypt-map}'s location than Fig. \ref{fig:sunset-map}'s since the probability distribution is more concentrated on a specific area. Thus, the estimation of the former can be considered more reliable than the latter. The spatial distribution of cells is essential information for the geolocation estimation problem, differentiating it from the general image classification task. Hence, our goal in this paper is to exploit this information to improve the reliability of the model's predictions.

\subsection{Localizability}
\label{sec:localizability}
To develop and evaluate a methodology for image localizability, we have to associate all images in a dataset with ground-truth labels that indicate localizability. Moreover, labeling images as localizable or not is highly subjective and depends on the collection of images recognized by the prospective annotator. To address the former issue, we define localizability at a certain scale $d$ (distance tolerance from the ground truth location). To address the latter issue, we approximate localizability in terms of our model's ability to infer location from the input image, i.e. we assess which images our employed model is able to predict correctly. Therefore, all images that our model is able to predict within a certain distance from their ground-truth location are labeled as localizable, and all other images are labeled as non-localizable.

More formally, given a geolocation estimation model $f$, the localizability of an image $x \in \mathbb{R}^{M \times N \times 3}$ at distance $d$ is defined as:
\begin{equation}
\mathcal{L}_{f}(x; d) = \begin{cases}
                            1, & \text{if GCD}(f(x), y) < d \\
                            0, & \text{otherwise}
                        \end{cases}
\end{equation}
where GCD$(\cdot, \cdot)$ is the Great Circle Distance between two locations, and $y$ is the ground-truth location of $x$.

\subsection{Selective Prediction}
\label{sec:confidence}
Predicting which images are localizable according to our model's geolocation capability can be formulated as a selective prediction scheme following the formulation in \cite{foundations}. Here, we adapt their definitions to fit the needs of the geolocation estimation task. Our aim is to build a selective geolocation system $(f, g)$ such that $f$ is the base geolocation module, as described in section \ref{sec:geolocation-estimation}, and $g$ is the selection function. Then our selective geolocation system is defined as:

\begin{equation}
    (f, g) (x; d) =
    \begin{cases}
    f(x; d), & \text{if } g(x; d) = 1 \\
    \textit{abstain}, & \text{if } g(x; d) = 0
    \end{cases}
\end{equation}

The selection function $g$ is usually modeled based on a confidence function $\kappa_f$ (which measures our model's confidence or uncertainty), a scale $d$ and a tunable threshold $\theta$. For $\kappa_f$ measuring confidence, $g(x; d)$ is defined as follows:
\begin{equation}
    g(x; d) =
    \begin{cases}
    1, & \text{if } \kappa_f(x; d) \geq \theta \\
    0, & \text{if } \kappa_f(x; d) < \theta
    \end{cases}
\end{equation}

Let $P(X, Y)$ be the distribution over $\mathcal{X} \times \mathcal{Y}$, where $\mathcal{X}$ is the image space and $\mathcal{Y}$ the coordinate space,  characterizing the probability of image $X$ being captured at geographical coordinates $Y$. Given an underlying distribution $P$, a confidence function $\kappa_f$ and a scale $d$, varying the parameter $\theta$ determines the performance of our selective geolocation system, which can be expressed using coverage and risk, as follows:

\textbf{Coverage} is the mass probability of the non-rejected region in $\mathcal{X}$, and can be approximated given enough i.i.d. samples $(x_i, y_i)$ from $P$ as follows:

\begin{equation}
    \phi_d(f, g) \triangleq \mathbb{E}_{P(X)} \left[ g(x; d) \right] \approx \dfrac{1}{N} \sum_{i=1}^{N} g(x_i; d)
\end{equation}

\textbf{Risk} is the expected percentage of the kept images that will be predicted outside a radius $d$, and can be approximated given enough i.i.d. samples $(x_i, y_i)$ from $P$ as follows:
\begin{equation}
    R_d(f, g) \triangleq \mathbb{E}_{P(X, Y)} \left[ l_d(f(x), y) \right] \approx \dfrac{ \frac{1}{N} \sum_{i=1}^{N} l_d(f(x_i), y_i) g(x_i; d) }{\phi_d(f, g)}
\end{equation}
where $l_d$ is a loss function defined as:
\begin{equation}
    l_d(l_1, l_2) =
    \begin{cases}
        1, &\text{if GCD}(l_1, l_2) > d \\
        0, &\text{if GCD}(l_1, l_2) \leq d
    \end{cases}
\end{equation}

\subsection{Estimating Image Localizability}
\label{sec:estimating-image-localizability}
Our main goal in this work is to find good confidence functions and thresholds for the function $g$. For $f$, we employ a geolocation system similar to \cite{planet}; however, any geolocation system that tackles geolocation as a classification problem can be used.

\subsubsection{Spatial Entropy (SE)} measures the dispersion of cell probabilities around the globe. To calculate the $\text{SE}(p; d)$ of the cell distribution $\{ p \} _{i=0}^{|C|} $ at scale $d$, we initially select the most probable cell and merge all cells within distance $d$ from it to form a super-cell. The probability of the super-cell derives from the sum of the cell probabilities of the individual cells. Then, we ignore all cells merged to the super-cell and find the next most probable cell from the remaining ones. Similarly, we merge it with its neighboring cells that are inside a radius $d$. This process is repeated until the cumulative probability of the super-cells accounts for the 90\% of the total confidence\footnote{We empirically found this to remove noise from cell distributions compared to considering cells accounting for 100\% of the total confidence.}. We denote as $\{ \bar{p} \} _{i=0}^{|C'|} $ the new probability distribution of the super-cells; hence, SE is defined as:
\begin{equation}
    \text{SE}(p; d) = - \sum_{i=0}^{|C'|} \bar{p}_i \log_2 \bar{p}_i
\end{equation}

Higher Spatial Entropy indicates lower confidence; therefore, we devise the selection function $g_{SE}$ as:
\begin{equation}
    g_{SE}(x) =
    \begin{cases}
    1, & \text{if } \text{SE}(p; d) \leq \theta_{SE} \\
    0, & \text{if } \text{SE}(p; d) > \theta_{SE}
    \end{cases}
\end{equation}
where $\theta_{SE}$ is a tunable threshold.

\subsubsection{Prediction Density (PD)} measures the concentration of cell probability in a particular region instead of its dispersion around the globe. To calculate the $\text{PD}(p; d)$ of the cell distribution $\{ p \} _{i=0}^{|C|} $ at scale $d$, we accumulate the model's cell probabilities in a radius $d$ around the most probable cell, which can be considered as the model's confidence that an input image can be localized at scale $d$. This is formulated as follows:
\begin{equation}
    \text{PD}(p; d) = \sum_{c \in C} p_c \cdot \mathbbm{1} \left[ \text{GCD}(c, \argmax_{c' \in C} p_{c'}) \leq d \right]
\end{equation}
where $\mathbbm{1}$ is the indicator function. Higher Prediction Density denotes higher confidence; therefore, we devise the selection function $g_{PD}$ as:
\begin{equation}
    g_{PD}(x) =
    \begin{cases}
    1, & \text{if } \text{PD}(p; d) \geq \theta_{PD} \\
    0, & \text{if } \text{PD}(p; d) < \theta_{PD}
    \end{cases}
\end{equation}
where $\theta_{PD}$ is a tunable threshold.

\section{Evaluation Set-up}
\label{sec:eval-setup}

\subsection{Datasets}
To train our geolocation model, we use the training split of the MediaEval Placing Task 2016 dataset (\textit{MP-16 train}) \cite{mp16}, which is a subset of the Yahoo Flickr Creative Commons 100 Million (\textit{YFCC100M}) \cite{yfcc100m}. It consists of 4,723,695 images posted on Flickr with their metadata, among which geographical coordinates. We also use the YFCC25k dataset from \cite{hierarchical}, composed of 25,600 randomly selected images from YFCC100M (excluding images from MP-16 train), for validation. Due to the unavailability of several images, we end up with a total of 23,007 images. Finally, for evaluation, we use the Im2GPS \cite{im2gps} and Im2GPS3k \cite{im2gps3k} datasets, provided by the original authors, consisting of 237 and 3,000 images, respectively.

\subsection{Implementation Details}
\label{sec:implementation-details}
For the cell partitioning described, we adopt the fine partitioning from \cite{hierarchical} and terminate cell splitting when each cell contains between 50 and 1,000 images from the MP-16 train. We discard all cells that end up with less than 50 photos. This results in 13,662 cells and 4,071,346 images for training. Although the particular partitioning implementation could affect the geolocation estimation performance, we are primarily interested in the performance of our localizability methods, and hence we do not consider alternate partitionings.

For the geolocation model $f$, we use EfficientNet-B4 \cite{efficientnet} as our backbone CNN and replace its last layer with a linear layer consisting of 13,662 neurons corresponding to our total number of cells. We replicate the pre-processing and training pipeline of Kordopatis et al. \cite{kordopatis-geolocation}, and we do not use any further additions such as hierarchical partitioning \cite{hierarchical} or the MvMF loss \cite{mvmf}.

\subsection{Competing Approaches}
\label{sec:rivals}
In Section \ref{sec:experiments}, we compare our selection functions against two baseline runs (which serve to visualize selective performance limits) and two state-of-the-art methods in selective prediction, briefly described below:

\textbf{Random} selection function: randomly selects whether to predict $f(x)$ or abstain from providing a prediction, with an equal probability of 50\%.

\textbf{Ideal} selection function: selects images based on their ground-truth localizability values, prioritizing the images considered localizable.

\textbf{Softmax Response (SR)}~\cite{baseline-softmax}: uses the maximum probability after the final softmax layer, i.e. the maximum cell probability, as confidence function. It has been employed for selective prediction in \cite{selective-prediction}. Note that this selection function cannot be intrinsically adapted for the different scales $d$.

\textbf{Monte-Carlo Dropout (MC)}~\cite{mc-dropout}: uses as confidence function the variance of the softmax response of multiple forward passes of an input image with dropout applied in the final layer. This is shown to be a good approximation of a Bayesian Neural Network with Gaussian parameter priors \cite{mc-dropout} and a state-of-the-art method in selective prediction \cite{selective-prediction}. Following \cite{selective-prediction} we use a dropout of 0.5. Note that again this selection function cannot be intrinsically adapted for the different scales $d$.

\section{Experiments}
\label{sec:experiments}

\begin{figure}[t]
    \includegraphics[width=0.98\linewidth]{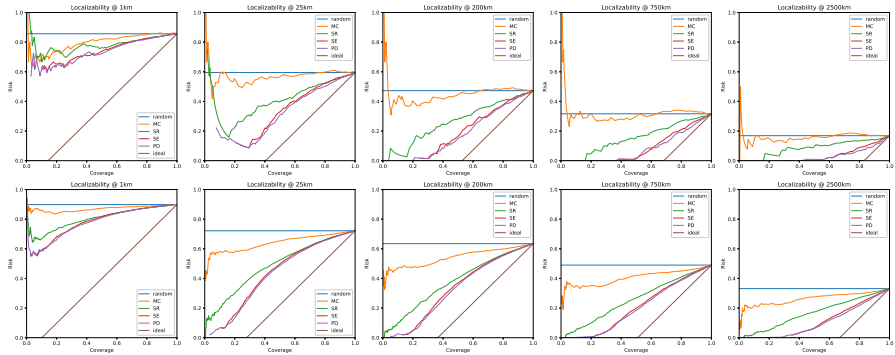}
    \caption{Risk-Coverage curves for six selection functions on Im2GPS (top row) and Im2GPS3k (bottom row). Lower is better.}
    \label{fig:rc}
\end{figure}

\subsection{Selective Geolocation Performance}
\label{sec:exp-1}
We benchmark the selective prediction performance of the proposed selection functions $g_{SE}$ and $g_{PD}$ against the competing approaches. We evaluate them in both Im2GPS and Im2GPS3k at scales $d = $ 1, 25, 200, 750 and 2500km, which are the most widely reported scales and correspond to street, city, region, country and continent level granularity scales.

First, we present the Risk-Coverage (RC) curves for each dataset, illustrated in Fig. \ref{fig:rc}. These curves are obtained by computing the risk and coverage of each method for different values of the threshold $\theta$. Both  our selection functions achieve state-of-the-art performance, yielding lower risk at every coverage level in all datasets and scales. SE and PD perform similarly, with PD consistently outperforming SE by a small margin. Moreover, in coarser granularity scales, the performance gap between our selection functions and the competing SR and MC widens considerably, with our selection functions reaching close to the ideal. This can probably be attributed to their intrinsic adaptation to different scales. Finally, it is evident that RC curves on Im2GPS3k are smoother and more monotonous, which is expected due to its greater size and variety of images.

\begin{table}[t]
\begin{subtable}{0.3\textwidth}
\scriptsize
\centering
\begin{tabular}{l|c|c|c|c|}
\cline{2-5}
                                  & \textbf{Acc $\uparrow$} & \textbf{F1 $\uparrow$} & \textbf{OR $\downarrow$} & \textbf{OC $\uparrow$} \\ \hline
\multicolumn{1}{|l|}{\textbf{SR}} & 74\%                    & 38\%                   & 71\%                     & 27\%                   \\ 
\multicolumn{1}{|l|}{\textbf{MC}} & \textbf{83\%}           & 23\%                   & 67\%                     & 5\%                    \\ \hline
\multicolumn{1}{|l|}{\textbf{SE}} & 76\%                    & 44\%                   & 67\%                     & \textbf{28\%}          \\ 
\multicolumn{1}{|l|}{\textbf{PD}} & 78\%                    & \textbf{47\%}          & \textbf{64\%}            & 27\%                   \\ \hline
\end{tabular}
\caption{$d = 1$ km}
\end{subtable}
\quad
\begin{subtable}{0.3\textwidth}
\scriptsize
\centering
\begin{tabular}{l|c|c|c|c|}
\cline{2-5}
                                  & \textbf{Acc $\uparrow$} & \textbf{F1 $\uparrow$} & \textbf{OR $\downarrow$} & \textbf{OC $\uparrow$} \\ \hline
\multicolumn{1}{|l|}{\textbf{SR}} & 70\%                    & 66\%                   & 40\%                     & 49\%                   \\ 
\multicolumn{1}{|l|}{\textbf{MC}} & 62\%                    & 31\%                   & 53\%                     & 19\%                   \\ \hline
\multicolumn{1}{|l|}{\textbf{SE}} & 73\%                    & 72\%                   & 37\%                     & \textbf{55\%}          \\ 
\multicolumn{1}{|l|}{\textbf{PD}} & \textbf{79\%}           & \textbf{77\%}          & \textbf{31\%}            & 50\%                   \\ \hline
\end{tabular}
\caption{$d = 25$ km}
\end{subtable}
\quad
\begin{subtable}{0.3\textwidth}
\scriptsize
\centering
\begin{tabular}{l|c|c|c|c|}
\cline{2-5}
                                  & \textbf{Acc $\uparrow$} & \textbf{F1 $\uparrow$} & \textbf{OR $\downarrow$} & \textbf{OC $\uparrow$} \\ \hline
\multicolumn{1}{|l|}{\textbf{SR}} & 70\%                    & 73\%                   & 31\%                     & 57\%                   \\ 
\multicolumn{1}{|l|}{\textbf{MC}} & 53\%                    & 36\%                   & 38\%                     & 25\%                   \\ \hline
\multicolumn{1}{|l|}{\textbf{SE}} & 76\%                    & 80\%                   & 29\%                     & \textbf{67\%}          \\ 
\multicolumn{1}{|l|}{\textbf{PD}} & \textbf{78\%}           & \textbf{81\%}          & \textbf{26\%}            & 64\%                   \\ \hline
\end{tabular}
\caption{$d = 200$ km}
\end{subtable}

\begin{subtable}{0.33\textwidth}
\scriptsize
\centering
\begin{tabular}{l|c|c|c|c|}
\cline{2-5}
                                  & \textbf{Acc $\uparrow$} & \textbf{F1 $\uparrow$} & \textbf{OR $\downarrow$} & \textbf{OC $\uparrow$} \\ \hline
\multicolumn{1}{|l|}{\textbf{SR}} & 70\%                    & 79\%                   & 24\%                     & 74\%                   \\ 
\multicolumn{1}{|l|}{\textbf{MC}} & 49\%                    & 53\%                   & 28\%                     & 41\%                   \\ \hline
\multicolumn{1}{|l|}{\textbf{SE}} & 79\%                    & 86\%                   & 20\%                     & \textbf{78\%}                   \\ 

\multicolumn{1}{|l|}{\textbf{PD}} & \textbf{84\%}           & \textbf{88\%}          & \textbf{12\%}            & 70\%          \\ \hline
\end{tabular}
\caption{$d = 750$ km}
\end{subtable}
\quad
\centering
\begin{subtable}{0.33\textwidth}
\scriptsize
\centering
\begin{tabular}{l|c|c|c|c|}
\cline{2-5}
                                  & \textbf{Acc $\uparrow$} & \textbf{F1 $\uparrow$} & \textbf{OR $\downarrow$} & \textbf{OC $\uparrow$} \\ \hline
\multicolumn{1}{|l|}{\textbf{SR}} & 82\%                    & 90\%                   & 14\%                     & \textbf{88\%}          \\ 
\multicolumn{1}{|l|}{\textbf{MC}} & 61\%                    & 74\%                   & 16\%                     & 62\%                   \\ \hline
\multicolumn{1}{|l|}{\textbf{SE}} & 85\%                    & 91\%                   & 10\%                     & 85\%                   \\ 
\multicolumn{1}{|l|}{\textbf{PD}} & \textbf{87\%}           & \textbf{92\%}          & \textbf{8\%}             & 85\%                   \\ \hline
\end{tabular}
\caption{$d = 2500$ km}
\end{subtable}

\caption{Selective prediction performance of our two selection functions against state-of-the-art on Im2GPS. We report the localizability accuracy and the F1-score as well as the optimal risk (OR) and coverage (OC) (for more details read Section \ref{sec:exp-1}). $\uparrow$ indicates that higher is better, and $\downarrow$ that lower is better.}
\label{tab:localizability-scores-im2gps}
\end{table}

\begin{table}[t]
\begin{subtable}{0.3\textwidth}
\scriptsize
\centering
\begin{tabular}{l|c|c|c|c|}
\cline{2-5}
                                         & \textbf{Acc $\uparrow$} & \textbf{F1 $\uparrow$} & \textbf{OR $\downarrow$} & \textbf{OC $\uparrow$} \\ \hline
\multicolumn{1}{|l|}{\textbf{SR}}        & 84\%                    & 34\%                   & 70\%                   & 14\%                     \\ 
\multicolumn{1}{|l|}{\textbf{MC}}        & \textbf{87\%}           & 7\%                    & 84\%                   & 3\%                      \\ \hline
\multicolumn{1}{|l|}{\textbf{SE}}        & 85\%                    & 44\%                   & 64\%                   &  \textbf{16\%}           \\ 
\multicolumn{1}{|l|}{\textbf{PD}}        & 86\%                    & \textbf{45\%}          & \textbf{62\%}          &  15\%                    \\ \hline
\end{tabular}
\caption{$d = 1$ km}
\end{subtable}
\quad
\begin{subtable}{0.3\textwidth}
\scriptsize
\centering
\begin{tabular}{l|c|c|c|c|}
\cline{2-5}
                                         & \textbf{Acc $\uparrow$} & \textbf{F1 $\uparrow$} & \textbf{OR $\downarrow$} & \textbf{OC $\uparrow$} \\ \hline
\multicolumn{1}{|l|}{\textbf{SR}}        & 78\%                    & 62\%                   & 39\%                   & 29\%                     \\ 
\multicolumn{1}{|l|}{\textbf{MC}}        & 70\%                    & 23\%                   & 58\%                   & 11\%                     \\ \hline
\multicolumn{1}{|l|}{\textbf{SE}}        & 83\%                    & 72\%                   & 35\%                   & \textbf{34\%}            \\ 
\multicolumn{1}{|l|}{\textbf{PD}}        & \textbf{85\%}           & \textbf{74\%}          & \textbf{29\%}          & 31\%                     \\ \hline
\end{tabular}
\caption{$d = 25$ km}
\end{subtable}
\quad
\begin{subtable}{0.3\textwidth}
\scriptsize
\centering
\begin{tabular}{l|c|c|c|c|}
\cline{2-5}
                                         & \textbf{Acc $\uparrow$} & \textbf{F1 $\uparrow$} & \textbf{OR $\downarrow$} & \textbf{OC $\uparrow$} \\ \hline
\multicolumn{1}{|l|}{\textbf{SR}}        & 76\%                    & 69\%                   & 33\%                   & 38\%                     \\ 
\multicolumn{1}{|l|}{\textbf{MC}}        & 65\%                    & 35\%                   & 48\%                   & 18\%                     \\ \hline
\multicolumn{1}{|l|}{\textbf{SE}}        & 80\%                    & 75\%                   & 31\%                   & \textbf{43\%}            \\ 
\multicolumn{1}{|l|}{\textbf{PD}}        & \textbf{82\%}           & \textbf{77\%}          & \textbf{26\%}          & 40\%                     \\ \hline
\end{tabular}
\caption{$d = 200$ km}
\end{subtable}

\begin{subtable}{0.33\textwidth}
\scriptsize
\centering
\begin{tabular}{l|c|c|c|c|}
\cline{2-5}
                                         & \textbf{Acc $\uparrow$} & \textbf{F1 $\uparrow$} & \textbf{OR $\downarrow$} & \textbf{OC $\uparrow$} \\ \hline
\multicolumn{1}{|l|}{\textbf{SR}}        & 72\%                    & 74\%                   & 28\%                   & 53\%                     \\ 
\multicolumn{1}{|l|}{\textbf{MC}}        & 57\%                    & 50\%                   & 37\%                   & 34\%                     \\ \hline
\multicolumn{1}{|l|}{\textbf{SE}}        & 78\%                    & 80\%                   & 24\%                   & \textbf{56\%}            \\ 
\multicolumn{1}{|l|}{\textbf{PD}}        & \textbf{80\%}           & \textbf{81\%}          & \textbf{22\%}          & 55\%                     \\ \hline
\end{tabular}
\caption{$d = 750$ km}
\end{subtable}
\quad
\centering
\begin{subtable}{0.33\textwidth}
\scriptsize
\centering
\begin{tabular}{l|c|c|c|c|}
\cline{2-5}
                                         & \textbf{Acc $\uparrow$} & \textbf{F1 $\uparrow$} & \textbf{OR $\downarrow$} & \textbf{OC $\uparrow$} \\ \hline
\multicolumn{1}{|l|}{\textbf{SR}}        & 70\%                    & 79\%                   & 25\%                   & \textbf{73\%}            \\ 
\multicolumn{1}{|l|}{\textbf{MC}}        & 58\%                    & 65\%                   & 28\%                   & 53\%                     \\ \hline
\multicolumn{1}{|l|}{\textbf{SE}}        & 78\%                    & 84\%                   & 18\%                   & 71\%                     \\ 
\multicolumn{1}{|l|}{\textbf{PD}}        & \textbf{80\%}           & \textbf{86\%}          & \textbf{16\%}          & 70\%                     \\ \hline
\end{tabular}
\caption{$d = 2500$ km}
\end{subtable}

\caption{Selective prediction performance of our two selection functions against state-of-the-art on Im2GPS3k. We report the localizability accuracy and the F1-score as well as the optimal risk (OR) and coverage (OC) (for more details read Section \ref{sec:exp-1}). $\uparrow$ indicates that higher is better, and $\downarrow$ that lower is better.}
\label{tab:localizability-scores-im2gps3k}
\end{table}

Although risk-coverage curves give a comprehensive insight of the selective prediction performance, we need to determine a specific threshold $\theta$ that separates localizable and non-localizable images given a selection function. To do so, we chose the $\theta$ value that corresponds to the coverage that equals the percentage of images $f$ can successfully localize. We learn this value on the validation YFCC25K dataset for each selection function and each granularity scale. We call the risk and coverage at this threshold \textit{Optimal Risk} (OR) and \textit{Optimal Coverage} (OC) respectively. We also report the classification accuracy and the F1-score of the positive class.

Tables \ref{tab:localizability-scores-im2gps} and \ref{tab:localizability-scores-im2gps3k} display the results of the selection functions on the two evaluation datasets. We note that high accuracy in finer scales $d$ is not indicative of good separation between localizable and non-localizable images due to the class imbalance; however, combined with F1-score, they provide useful insights. In particular, it is evident that the proposed SE and PD achieve better class separation than the SR and MC, with PD slightly surpassing SE. Moreover, in most cases, the selected threshold $\theta$ for our methods leads to lower risk and wider coverage compared to the competition.

\subsection{Selective Geolocation Reliability}
We present a quantitative and qualitative assessment of the performance of our selective models $(f, g_{SE})$ and $(f, g_{PD})$ compared to $f$. 

\begin{table}[t]
\centering
\hspace{-1cm}
\resizebox{0.43\textwidth}{!}{
\begin{subtable}[h]{0.5\linewidth}
    \centering
    \begin{tabular}{l|ccccc|}
    \cline{2-6}
    \textbf{}                                           & \textbf{1km} & \textbf{25km} & \textbf{200km} & \textbf{750km} & \textbf{2500km} \\ \hline
    \multicolumn{1}{|l|}{$f$}                   & 14.3         & 40.5          & 52.7           & 68.3           & 83.1            \\ \hline
    \multicolumn{1}{|l|}{$(f, g_{SE})^L$}     & 24.4         & 62.6         & 74.0           & 86.2           & 91.6            \\
    \multicolumn{1}{|l|}{$(f, g_{PD})^L$}     & 27.1         & 69.5          & 79.6           & 89.8          & 94.0            \\ \hline
    \multicolumn{1}{|l|}{$(f, g_{SE})^N$}     & 1.9          & 13.2          & 26.4           & 46.2           & 72.6            \\
    \multicolumn{1}{|l|}{$(f, g_{PD})^N$}     & 1.6         & 11.7          & 26.0           & 47.0           & 72.6            \\ \hline
    \end{tabular}
    \caption{Im2GPS}
\end{subtable}}
\hspace{0.6cm}
\resizebox{0.43\textwidth}{!}{
\begin{subtable}{0.5\linewidth}
    \centering
    \begin{tabular}{r|ccccc|}
    \cline{2-6}
    \textbf{}                                           & \textbf{1km} & \textbf{25km} & \textbf{200km} & \textbf{750km} & \textbf{2500km} \\ \hline
    \multicolumn{1}{|l|}{$f$}               & 10.1         & 27.8          & 36.5           & 51.0             & 66.8            \\ \hline
    \multicolumn{1}{|l|}{$(f, g_{SE})^L$}      & 24.6         & 65.3          & 77.4           & 86.9           & 92.6            \\
    \multicolumn{1}{|l|}{$(f, g_{PD})^L$} & 26.6         & 70.5          & 80.8           & 89.1           & 94.1            \\ \hline
    \multicolumn{1}{|l|}{$(f, g_{SE})^N$}      & 2.7            & 8.4          & 17.7           & 35.0             & 55.6            \\
    \multicolumn{1}{|l|}{$(f, g_{PD})^N$} & 2.6          & 8.2           & 15.3           & 32.3           & 54.5            \\ \hline
    \end{tabular}
    \caption{Im2GPS3k}
\end{subtable}}

\caption{Geolocation accuracies (\%) when evaluating the whole dataset and the Localizable (L) and Non-Localizable (N) subsets. The percentage of images on (L) corresponds to the Optimal Coverage (OC) column of Tables \ref{tab:localizability-scores-im2gps} and \ref{tab:localizability-scores-im2gps3k}.}
\label{tab:geolocation-street-level}
\end{table}

We split both Im2GPS and Im2GPS3k into a localizable and a non-localizable subset using our selective models at city-scale. Table \ref{tab:geolocation-street-level} displays the geolocation accuracies on these splits at all granularity scales, compared to the performance of the base model $f$ without a selection scheme. For fine and medium granularity scales, our selective models achieve more than double the geolocation accuracy of the base model $f$, with only a tiny portion of localizable images rejected by our functions. In particular, prediction density increased the geolocation accuracy on Im2GPS3k from 27.8\% to 70.5\% by discarding non-localizable images, from which only 8.2\% could have been successfully localized. This highlights the reliability current image geolocation models can achieve using the selective prediction mechanisms presented.

\begin{figure}[t]
    \centering
    \begin{subfigure}[h]{\linewidth}
        \centering
        \includegraphics[width=0.91\linewidth]{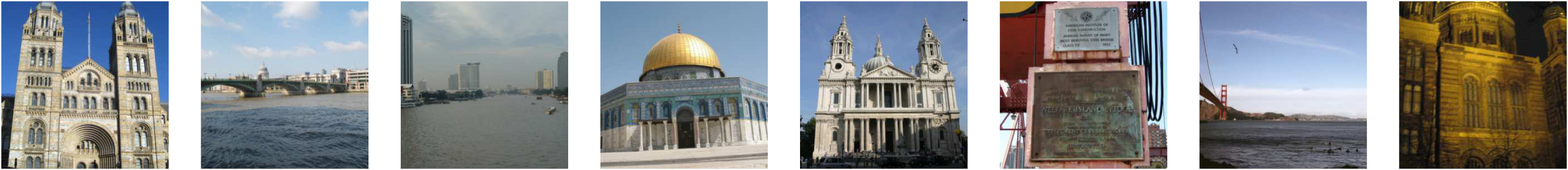}
        \caption{True positive: samples correctly predicted as localizable}
    \end{subfigure}
    \begin{subfigure}[h]{\linewidth}
        \centering
        \includegraphics[width=0.91\linewidth]{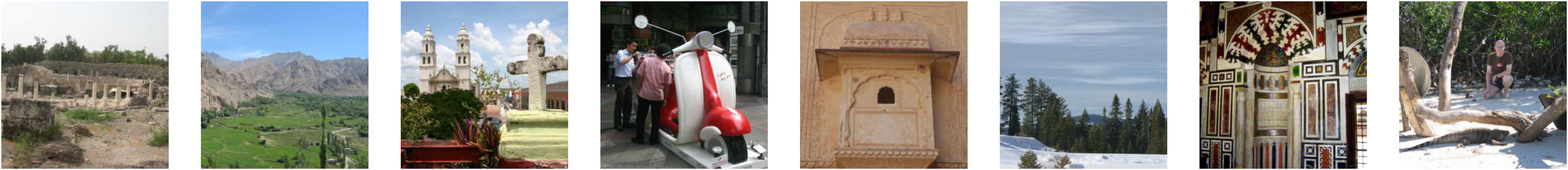}
        \caption{False positive: samples wrongly predicted as localizable}
    \end{subfigure}
    \begin{subfigure}[h]{\linewidth}
        \centering
        \includegraphics[width=0.91\linewidth]{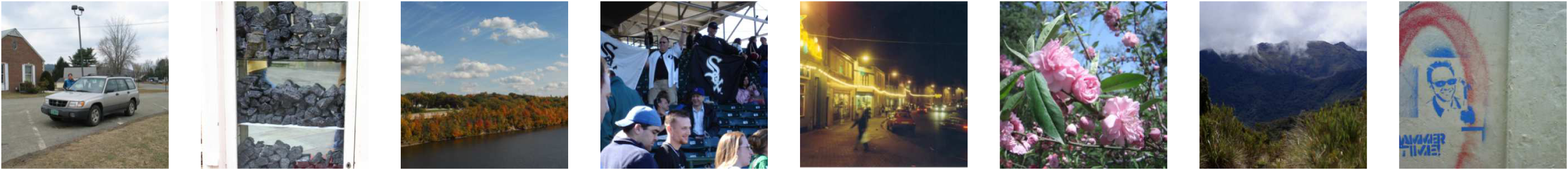}
        \caption{True negative: samples correctly predicted as non-localizable}
    \end{subfigure}
    \begin{subfigure}[h]{\linewidth}
        \centering
        \includegraphics[width=0.91\linewidth]{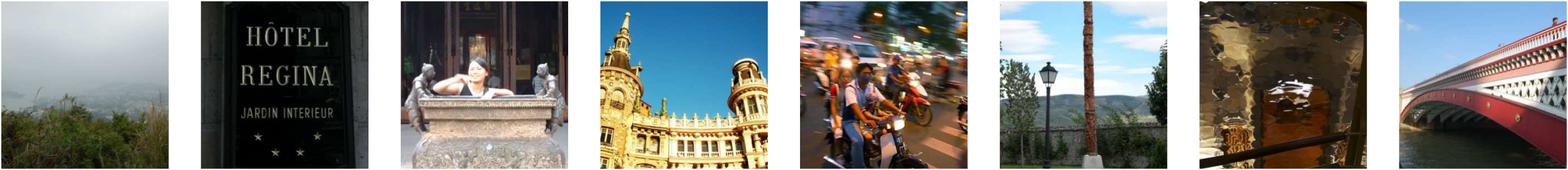}
        \caption{False negative: samples wrongly predicted as non-localizable}
    \end{subfigure}
    
    \caption{Sample predictions of Prediction Density (PD) on randomly selected images of the Im2GPS3k dataset.}
    \label{fig:samples}
\end{figure}

Fig. \ref{fig:samples} depicts image samples randomly selected from Im2GPS3k for qualitative evaluation of our methodology. Images are grouped by their predicted and ground-truth localizability in (a) true positive, (b) false positive, (c) true negative and (d) false negative, using PD at city-scale as the selection function. True positive samples either depict landmarks or characteristic elements that hint at very specific locations (e.g. the Golden Gate Bridge). True negative samples contain mostly generic scenes that should not even be attempted to be localized. The two images with the car and the lights could have been localized if they were in our training dataset, but even in that case, a similar scene can easily exist in multiple cities. False positive samples contain enough visual cues to be worthy of geolocation, however not enough for the required granularity. Finally, all false negative samples besides the Edificio Meneses picture are not localizable and their correct geolocation by our geolocation model could be attributed to presence of very similar images in the train dataset.

\section{Conclusions}
In this paper, we introduced the problem of image localizability detection and used it as a foundation for reliable image geolocation. We adapted a selective prediction methodology to the context of geolocation and presented two novel selection functions, Spatial Entropy and Prediction Density, tailored to the needs of the geolocation task. Our functions achieved superior selective performance compared to state-of-the-art on the two widely-used evaluation datasets. We also demonstrated how they can be exploited to abstain from geolocating non-localizable images, significantly boosting the geolocation performance in all granularity scales, and thus making current geolocation models more reliable. In the future, we plan to explore the design and evaluation of more sophisticated and trainable selection functions.

\bigskip
\noindent\textbf{Acknowledgments}:
This work has been supported by the projects WeVerify and MediaVerse, partially funded by the European Commission under contract number 825297 and 957252, respectively.


\end{document}